\documentclass{article}

\usepackage[final, nonatbib]{neurips_2022}

\usepackage[utf8]{inputenc} 
\usepackage[T1]{fontenc}    
\usepackage{hyperref}       
\usepackage{url}            
\usepackage{booktabs}       
\usepackage{amsfonts}       
\usepackage{nicefrac}       
\usepackage{microtype}      
\usepackage{xcolor}         

\usepackage[noend]{algorithm2e}
\RestyleAlgo{ruled}
\SetKwComment{Comment}{/* }{ */}
\usepackage{xspace}
\usepackage{makecell}
\usepackage{subcaption}
\usepackage{multirow}
\usepackage{threeparttable}
\usepackage{amsmath}
\usepackage{graphicx}
\usepackage{comment}

\newcommand{\newparagraph}[1]{\vspace{2pt}\noindent\textbf{#1\xspace~~}}

\newcommand{\slicefinder}{{SliceFinder}}
\newcommand{\sliceline}{{SliceLine}}
\newcommand{\autoslicer}{{AutoSlicer}}
\newcommand{\tf}{{TensorFlow}}

\title{AutoSlicer: Scalable Automated Data Slicing for ML Model Analysis}

\author{%
  Zifan Liu\thanks{Work done while interning at Google.} \\
  University of Wisconsin-Madison\\
  \texttt{zliu676@wisc.edu} \\
  \And
  Evan Rosen \\
  Google Inc. \\
  \texttt{embr@google.com} \\
  \And
  Paul Suganthan G. C. \\
  Google Inc. \\
  \texttt{paulgc@google.com} \\
}

\begin{document}

\maketitle

\begin{abstract}
Automated slicing aims to identify subsets of evaluation data where a trained model performs anomalously. This is an important problem for machine learning pipelines in production since it plays a key role in model debugging and comparison, as well as the diagnosis of fairness issues. Scalability has become a critical requirement for any automated slicing system due to the large search space of possible slices and the growing scale of data. We present \autoslicer, a scalable system that searches for problematic slices through distributed metric computation and hypothesis testing. We develop an efficient strategy that reduces the search space through pruning and prioritization. In the experiments, we show that our search strategy finds most of the anomalous slices by inspecting a small portion of the search space.
\end{abstract}

\section{Introduction}

In production settings, machine learning (ML) practitioners have to consider whether the input data have errors, whether a new version is good enough to replace the model currently used by the downstream stack, how to debug and improve model performance etc. Those require reliable model evaluation that can aid with validating, debugging and improving model performance. 

\textbf{Example 1:} \textit{Consider a continuous ML pipeline that trains on new data arriving in batches every day, to obtain a fresh model that needs to be pushed to the serving infrastructure. The ML engineer needs to validate that the new model does not perform worse than the previous one.}


Model metrics computed on the whole evaluation dataset can mask interesting or significant deviations of the same metrics computed on data slices that correspond to meaningful sub-populations. Thus, it is often desired to identify slices of data where the model performs poorly.

\textbf{Example 2:} \textit{Consider the same pipeline in Example 1. To ensure fairness and model quality, we want to avoid pushing the new model to serving if it performs poorly on sensitive data slices.}

\textbf{Example 3:} \textit{Consider an ML engineer who is building the first model for a task. The engineer tries to iteratively debug and improve the model. Knowing on which data slices the current model is performing poorly would aid the engineer to take actions towards improving the accuracy on them.}


To this end, we focus on the problem of automated slicing, aiming at identifying problematic data slices automatically. As part of an ML platform in production, the automated slicing system should satisfy the following requirements: 
(\textit{Scalability}) The system should be able to operate on distributed clusters and the search method should be efficient, since modern ML datasets are often too large to fit in memory, and enumerating the number all possible slices is infeasible.
(\textit{Reliability}) The system should provide uncertainty measures that tell the user how reliable the results are, since some slices may appear problematic by chance due to sampling errors. 
(\textit{Generality}) The system should be general and flexible to support common types of models and metrics.

\slicefinder~\cite{chung2019automated} identifies problematic slices by lattice search and hypothesis testing.
However, \slicefinder\ focuses on the single-machine setting, failing to scale on large datasets. \sliceline~\cite{sagadeeva2021sliceline} addresses the scalability issues by formulating automated slicing as linear-algebra problems, and solving them through distributed matrix computing. However, \sliceline\ does not provide uncertainty measures for the computed metric. In addition, \slicefinder\ and \sliceline\ only support metrics that are mean values over individual examples (e.g., average loss, accuracy), while we want a general framework that supports all common metrics in ML evaluation (e.g., AUC, F1 score).  

In this paper we describe \autoslicer, a scalable and reliable solution to the automatic slicing problem.
Our technical contributions include:
1. A distributed computing framework that performs sliced evaluation efficiently for a general family of ML metrics (e.g., those provided by \tf).
2. Uncertainty estimation for the results of metric comparison by distributed hypothesis testing.
3. A search strategy that identifies most of the problematic slices without exhausting the search space.

\section{Problem Definition}
We assume a dataset $D = \{(x^{(1)}_F, y^{(1)}), (x^{(2)}_F, y^{(2)}), \dots, (x^{(N)}_F, y^{(N)})\}$ with $N$ examples, where $x^{(n)}_F$ is the $n^{\text{th}}$ example and $y^{(n)}$ is the corresponding ground truth label. $F = \{F_1, F_2, \dots, F_M\}$ is the set of features in each example. We also assume a model $h$ to be tested, and a metric $\psi(S, h)$ measuring how well the model makes predictions about the ground truth labels on a subset $S \subseteq D$.

A slice $S_P$ is a subset of $D$ that satisfies a conjunctive predicate $P = \bigwedge_i F_{m_i} \in V_i$, where $V_i$ is a set of values from the domain of feature $F_{m_i}$. 
For categorical features, each $V_i$ contains one value from the feature domain. If the domain of a feature is too large (it contains more than $J$ values), we put each of the $J$ most frequent values into one singleton set, and all the rest into one set.
For numerical values, each $V_i$ corresponds to a bin. 
We call each $F_{m_i} \in V_i$ a singleton predicate, and use $|P|$ to denote the number of singleton predicates in $P$ (cross size of $P$). We use $O$ to denote the predicate that is always true, and thereby $S_O$ is the overall slice that contains all examples.

Automated slicing aims to find slices with significantly higher or lower metric values, compared to the overall slice or the same slice under a baseline model. Formally, we seek $S_P$ such that $\Delta \psi_{S_P}$ is significantly greater or less than $0$, where $\Delta \psi_{S_P} = \psi(S_P, h)-\psi(S_O, h)$ in the former scenario, and $\Delta \psi_{S_P} = \psi(S_P, h)-\psi(S_P, h')$ ($h'$ is a baseline model) in the latter. We quantify significance by $p$-values from hypothesis testing. Any slice with $p$-value less than $\alpha$ is considered significant where $\alpha$ is a user-specified threshold.

\begin{figure}
    \centering
    \includegraphics[width=0.95\linewidth]{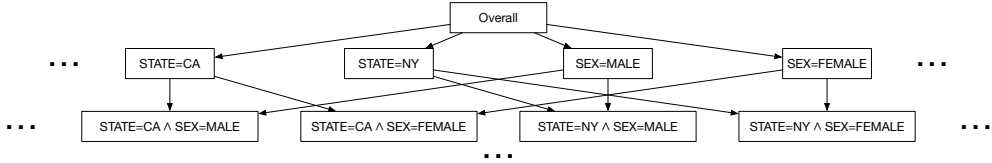}
    \vspace{-1mm}
    \caption{An example of the search space.}
    \label{fig:lattice}
    \vspace{-1mm}
\end{figure}

The search space forms a lattice structure, where the $l^\text{th}$ layer contains all possible slices with cross size $l$. An example of the search space is shown in Figure~\ref{fig:lattice}. 
To limit the number of candidate slices, we introduce the maximum cross size $L$ as a user-specified parameter. Another reason for setting the maximum cross size is that as the cross size increases, the results become less interpretable.

In some cases, the user may not want slices that are too small since they do not have a large impact on the overall performance of the model. 
Therefore, we also introduce the minimum slice size $N_\text{min}$ as another user-specified parameter.

In summary, we seek $S_P$'s such that the difference of a given metric $\Delta \psi_{S_P}$ is significantly greater (or less) than $0$, with $|P| \leq L$ and $|S_P| \geq N_\text{min}$, where $|S_P|$ is the number of examples in the slice.

\section{Distributed Computation Framework}
\autoslicer\ uses the Beam programming model \cite{beam} to express distributed programs that can be executed in a variety of environments.

\begin{figure*}
    \centering
    \includegraphics[width=0.84\linewidth]{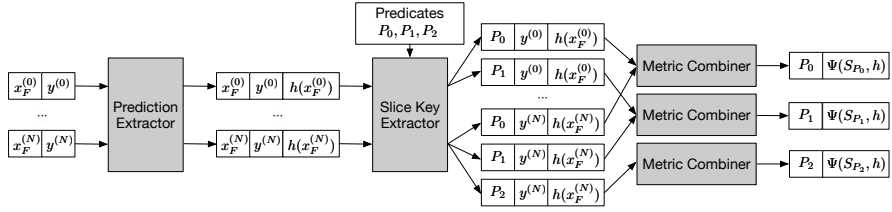}
    \caption{A diagram showing the distributed computation of sliced evaluations.}
    \label{fig:tfma-slicing}
\end{figure*}

\newparagraph{Sliced Evaluation}
\autoslicer\ is able to evaluate a wide variety of model performance metrics on slices of large datasets. In Figure~\ref{fig:tfma-slicing}, we show a diagram of the metric computing process.
First, the prediction extractor computes model predictions for each input example. Next, the slice key extractor checks a specified set of slicing predicates against the per-example features. The predicates are generated by the candidate slice generator described in Section~\ref{sec:strategy}.
For each matched example, a triplet containing the predicate, prediction and label is created.
Note that the prediction and the label for an example that matches more than one predicates will appear more than once in the output. 
Conversely, predicates that match no example will not be present in the output and therefore will not contribute to the computational costs of downstream processing.
Then the metric combiner aggregates the predictions and labels per predicate, which yields a collection of per-slice performance metrics. For example, to compute the precision of a binary classification model on a slice $S_P$, the combiner will count the number of true positives and false positives based on the predictions and labels whose keys match $P$. Finally, \autoslicer\ computes metric differences $\Delta \psi$ to yield another collection of per-slice differences. The differences are either between the metric on each slice and the entire dataset ($S_O$), or between the metrics on the same slice evaluated on the two models that need to be compared.

\newparagraph{Statistical Significance Testing}
In order to decide whether each metric difference $\Delta\psi$ is significant, \autoslicer\ performs distributed hypothesis testing. \autoslicer\ uses a parametric Poisson bootstrap approximation \cite{chamandy2012uncertainty} to estimate the sampling distribution of $\Delta\psi$. Each of the $B$ bootstrap replicates starts from a Poisson re-sampling of the entire input dataset, which is then fed to the sliced evaluation algorithm described above. This results in per-slice collections of $B$ metric differences, $\{\Delta\psi_b\}_{b=0}^{B}$, over which we compute the $t$-statistics for statistical significance testing where the null hypothesis is $\Delta\psi=0$. Per-slice $p$-values are then computed for significance-based output filtering.

\section{Search Strategies}\label{sec:strategy}
\autoslicer\ relies on a candidate slice generator that automatically proposes candidate slices from the search space. The candidate slice generator supports several search strategies. We describe them as follows. The complete algorithms for the last two are provided in Appendix~\ref{sec:alg}.

\newparagraph{Batch Strategy}
The candidate slice generator enumerates all the candidates at once, and then sends them to the slice key extractor for metric computation. 

\newparagraph{Iterative Strategy}
The candidate slice generator enumerates slices with cross size $l$ in the $l^\text{th}$ iteration. Metric computation is performed for the candidate slices in the current iteration before the beginning of the next. With this strategy, the search space is pruned based on the following criterion:
1. If a slice is significant, we do not consider any of its subsets as candidate slices in the subsequent iterations. This pruning rule follows \slicefinder~\cite{chung2019automated}.
2. If the size of a slice is less than the minimum slice size, we do not consider any of its subsets since the size can only be even smaller. 




\newparagraph{Priority Strategy}
Since the search space might be too large to exhaust, we want a strategy that finds most of the significant slices by inspecting only part of it.
In the priority strategy, we assign a priority score to each nonsignificant slice to prioritize those that are more likely to contain significant sub-slices for further exploration. We choose $p$-value as the priority score assuming that slices with lower $p$-values are more likely to contain significant sub-slices due to their more extreme metric values. This choice is validated in Appendix~\ref{sec:micro_benchmark}. 
In addition, we limit the number of candidate slices per iteration so that the computation cost will not be excessively high for any iteration. 

In the first iteration, all the candidates with cross size one are proposed. Afterwards, the nonsignificant slices are pushed into the priority queue. In each subsequent iteration, we pop base slices from the priority queue, and subdivide them by adding one more singleton predicate until the estimated number of nonempty candidates slices reaches the limit $K$. Then we compute the metrics and perform hypothesis testing for the candidates and push the new nonsignificant slices into the priority queue. In this strategy, we follow the same criterion as in the iterative strategy to prune the search space. 




Note that we use an estimated number of nonempty candidate slices to decide if enough base slices have been popped from the queue. This is because some candidates are empty, introducing no computation cost. 
With this estimation, we have a better control over the computation cost per iteration. 
We keep track of the nonempty rate for each cross size $l$, i.e., the ratio between the actual number of nonempty slices and the number of slice candidates with cross size $l$ that we have seen so far. When we count the number of nonempty slices, each candidate is discounted based on the nonempty rate corresponding to its cross size. For example, a candidate with cross size $3$ will be counted as $0.5$ slices if the nonempty rate of cross size $3$ is $0.5$. If we have not seen any slice with cross size $l$, the empty rate of cross size $l$ will be predicted as that of cross size $l-1$.

\section{Experimental Evaluation}

We compare the search strategies on a variety of real-world datasets. In the Appendix, we provide micro-benchmarks over several design choices and a scalability evaluation (Appendix~\ref{sec:micro_benchmark}, ~\ref{sec:scalability}).

We describe the datasets, ML models, and parameters of \autoslicer\ in Appendix~\ref{sec:exp_setup}. We run all the experiments on an internal cluster. Due to the resource allocation mechanism that we do not have control over, the wall time of the same run may vary a lot. Therefore, we report the total CPU time across workers for fair comparison of computation cost.

The comparison between the strategies is shown in Table~\ref{tab:strategy}. We run the priority strategy for 5 iterations, and set the target number of candidates per iteration ($K$) to be 12\% of the size of batch strategy search space. When the batch strategy cannot finish, we set $K$ to 2500.
Results under other configurations can be found in Appendix~\ref{sec:more_configs}.
The significant slices in this experiment are those the model performs poorly on compared to the overall dataset.
We report the number of significant slices found by each strategy (significant slices that are subsets of other significant ones are pruned), the actual number of candidate slices computed, the total CPU usage across workers and the amount of shuffle bytes. 
Note that due to the randomness of bootstrapping, the number of significant slices is slightly different across runs, which is the reason why the number of significant slices found by the priority strategy is larger than that by the exhaustive batch strategy in some cases.

The results show the effectiveness of our search space pruning and the priority strategy. On OpenML Electricity and UCI Census, the iterative strategy takes at least 45\% less CPU seconds and 90\% less shuffle bytes than the batch strategy due to the pruning. The priority strategy further reduces those costs. Compared to the iterative strategy, the priority strategy takes 40\% less CPU seconds on OpenML Electricity and 76\% less on UCI Census to find a comparable number of significant slices. The shuffle bytes are also reduced by at least 38\% on both datasets. For Safe Driving and Traffic Stop, the priority strategy is able to finish while the other two strategies run out of time.



\begin{table*}[h]
\caption{Comparison of different search strategies.}
\center
\label{tab:strategy}
\small
\setlength{\tabcolsep}{5pt}
\begin{tabular}{cccccc}
\toprule
Dataset & Strategy & \begin{tabular}{cc}\# Significant \\ Slices Found*\end{tabular} & \begin{tabular}{cc}\# Candidate \\ Slices\end{tabular}  & \begin{tabular}{cc} CPU Usage \\ (CPU-sec)\end{tabular} & \begin{tabular}{cc} Shuffle \\ Bytes (GB)\end{tabular}\\
\midrule
\multirow{3}{*}{\begin{tabular}{cc}OpenML \\ Electricity\end{tabular}} & Batch & 317 & 30000 & 168K & 233\\
                                    & Iterative & 327 & 23499 & 90.6K & 19.3 \\
                                    & Priority & 302 & 15114 & 54.9K & 9.68\\
\midrule
\multirow{3}{*}{\begin{tabular}{cc}UCI \\ Census\end{tabular}}  & Batch & 303 & 63861 & 412K & 971\\
                                    & Iterative & 316 & 29079 & 213K & 70.1\\
                                    & Priority & 374 & 30308 & 50.8K & 43.4\\
\midrule
\multirow{1}{*}{Traffic Stop**} 
                                    & Priority & 146 & 10157 & 1.34M & 326\\
\midrule                                    
\multirow{1}{*}{Safe Driving**} 
                                    & Priority & 369 & 10532 & 1.19M & 442\\
\bottomrule
\end{tabular}
\begin{tablenotes}
      \item * The number of significant slices may vary across runs due to the randomness of bootstrapping.
      \item ** The batch and the iterative strategy cannot finish in 8 hours. 
\end{tablenotes}
\end{table*} 




\clearpage

\bibliographystyle{acm}
\bibliography{references}

\appendix

\section{Appendix}

\subsection{Complete Algorithms for the Search Strategies}\label{sec:alg}
We show the complete algorithms for the iterative and priority strategies in Algorithm~\ref{alg:iterative} and Algorithm~\ref{alg:priority_iterative}.

\begin{algorithm}[hbt]
\caption{Slice finding algorithm with the iterative strategy}\label{alg:iterative}
\textbf{Inputs:} dataset $D$, model $h$, metric $\psi$, maximum cross size $L$, minimum slice size $N_\text{min}$, $p$-value threshold $\alpha$\;
\textbf{Outputs:} set of significant slices $\mathcal{S}$\;
$i \gets 1$ \Comment*[r]{Iteration number.}
$\mathcal{S} \gets \emptyset$ \Comment*[r]{Set of significant slices.}
$\mathcal{E} \gets \emptyset$ \Comment*[r]{Set of slices with size $< N_\text{min}$.}
$\mathcal{N}_\text{prev} \gets \{ S_O\}$ \Comment*[r]{Set of nonsignificant slices from the previous iteration. Only contain the overall slice in the beginning.}

$\mathcal{P}_\text{singleton} \gets \text{GetSetOfSingletonPredicates}(D)$\;
\While{$i \leq L$}{
    $\mathcal{C} \gets \emptyset$ \Comment*[r]{Set of candidate slices.}
    \For{$S_P \in \mathcal{N}_\text{prev}$}{
        \For{$P_\text{singleton} \in \mathcal{P}_\text{singleton}$}{
            $P' \gets P \land  P_\text{singleton}$\;
            \If{$S_{P'}$ is not a subset of any slices in $\mathcal{S}$ or $\mathcal{E}$}{
                $\mathcal{C} \gets \mathcal{C} \cup \{ S_{P'}\}$ \;
            }
        }
    }
    $\mathcal{S}_\text{new}, \mathcal{N}_\text{new} \gets \text{MetricComputingAndHypothesisTest}(\mathcal{C}, D, h, \psi, \alpha)$\; 
    $\mathcal{S}_\text{new}, \mathcal{N}_\text{new}, \mathcal{E}_\text{new}\gets \text{FilterBySize}(\mathcal{S}_\text{new}, \mathcal{N}_\text{new}, N_\text{min})$\;
    $\mathcal{S} \gets \mathcal{S} \cup \mathcal{S}_\text{new}$\;
    $\mathcal{E} \gets \mathcal{E} \cup \mathcal{E}_\text{new}$\;
    $\mathcal{N}_\text{prev} \gets \mathcal{N}_\text{new}$\;
    $i \gets i+1$\;
}
\end{algorithm}
\begin{algorithm}[hbt]
\caption{Slice finding algorithm with the priority strategy}\label{alg:priority_iterative}
\textbf{Inputs:} dataset $D$, model $h$, metric $\psi$, maximum cross size $L$, minimum slice size $N_\text{min}$, $p$-value threshold $\alpha$, number of iterations $I$, target number of candidate slices per iteration $K$\;
\textbf{Outputs:} set of significant slices $\mathcal{S}$\;
$i \gets 1$ \Comment*[r]{Iteration number.}
$\mathcal{S} \gets \emptyset$ \Comment*[r]{Set of significant slices.}
$\mathcal{E} \gets \emptyset$ \Comment*[r]{Set of slices with size $< N_\text{min}$.}
$\mathcal{Q} \gets []$ \Comment*[r]{Priority queue sorted by $p$-values.}

$\mathcal{P}_\text{singleton} \gets \text{GetSetOfSingletonPredicates}(D)$\;
\While{$i \leq I$}{
    $\mathcal{C} \gets \emptyset$ \Comment*[r]{Set of candidate slices.}
    \eIf{$l=1$}{
        \For{$P \in \mathcal{P}_\text{singleton}$}{
            $\mathcal{C} \gets \mathcal{C} \cup \{ S_P \}$
        }
    }{
        $k \gets 0$ \Comment*[r]{Estimated number of nonempty candidate slices.}
        \While{$k < K$}{
            $S_P \gets \text{PopFromQueue}(\mathcal{Q})$
            \For{$P_\text{singleton} \in \mathcal{P}_\text{singleton}$}{
                $P' \gets P \land  P_\text{singleton}$\;
                \If{$S_{P'}$ is not a subset of any slices in $\mathcal{S}$ or $\mathcal{E}$ and $|P'| \leq L$}{
                    $\mathcal{C} \gets \mathcal{C} \cup \{ S_{P'}\}$ \;
                    $k \gets k + \text{NonEmptyRate}(|P'|)$\;
                }
            }
        }
    }
    $\mathcal{S}_\text{new}, \mathcal{N}_\text{new} \gets \text{MetricComputingAndHypothesisTest}(\mathcal{C}, D, h, \psi, \alpha)$\;
    $\text{UpdateNonEmptyRate}(\mathcal{C}, \mathcal{S}_\text{new}, \mathcal{N}_\text{new})$\;
    $\mathcal{S}_\text{new}, \mathcal{N}_\text{new}, \mathcal{E}_\text{new}\gets \text{FilterBySize}(\mathcal{S}_\text{new}, \mathcal{N}_\text{new}, N_\text{min})$\;
    $\mathcal{S} \gets \mathcal{S} \cup \mathcal{S}_\text{new}$\;
    $\mathcal{E} \gets \mathcal{E} \cup \mathcal{E}_\text{new}$\;
    \For{$S_P \in \mathcal{N}_\text{new}$}{
        $\text{PushToQueue}(\mathcal{Q}, S_P)$
    }
    $i \gets i+1$\;
}
\end{algorithm}

\subsection{Additional Experimental Details}

\subsubsection{Experimental Setup}\label{sec:exp_setup}
\newparagraph{Datasets and Models}
We consider four real-world datasets that contain a mixture of numerical and categorical features. The properties of these datasets are shown in Table~\ref{tab:data_info}. The prediction task is binary classification and the metric is binary accuracy. We consider different types of models that are widely used in practice, including random forest (RF), deep neural network (DNN) and gradient-boosted decision tree (GBDT). The models for OpenML Electricity and UCI Census are RF and GBDT respectively; the model for the others is DNN. 
\begin{table}[t]
\caption{Properties of the datasets in the experiments.}
\center
\label{tab:data_info}
\small
\begin{tabular}{lccc}
\toprule
Dataset & \# Records & \# Numerical Features & \# Categorical Features \\
\midrule
OpenML Electricity~\cite{vanschoren2014openml} & 3,621 & 8 & 1\\
UCI Census~\cite{kohavi1996scaling}    & 3,918 & 5 & 9\\
Traffic Stop~\cite{pierson2020large}  & 72,398 & 9 & 7\\
Safe Driving~\cite{safedrivingdataset}    & 47,655 & 58 & 7\\
\bottomrule
\end{tabular}
\vspace{-2mm}
\end{table}

\newparagraph{Parameters}
In all the experiments, we set the maximum cross size $L$ to $3$, the minimum slice size $N_\text{min}$ to $1$, and the $p$-value threshold $\alpha$ to $0.01$ by default. We re-sample $20$ times for the Poisson bootstrap approximation. For categorical features, we set $J$ to $100$ so that the categories that are not among the top-$100$ frequent ones are combined in to one category. For numerical features, we bucketize the value domain so that each bin contains 10\% of the examples.

\subsubsection{More Configurations for the priority strategy}\label{sec:more_configs}
For the priority strategy, we fix the number of iterations to $5$, and set four different configurations (Priority-Config 1-4) by varying the target number of candidate slices per iteration ($K$). When the batch strategy can finish, we choose proper values for $K$ so that the sizes of the target search space are 10\%, 20\%, 40\%, 60\% of that of the batch strategy search space for Config 1-4. When the batch strategy cannot finish, we set $K$ to $2500$, $5000$, $10000$, $20000$ for Config 1-4 respectively. The results are shown in Table~\ref{tab:strategyA}.

\begin{table*}[ht]
\caption{Comparison of different search strategies.}
\vspace{-4mm}
\center
\label{tab:strategyA}
\small
\setlength{\tabcolsep}{5pt}
\begin{tabular}{cccccc}
\toprule
Dataset & Strategy & \begin{tabular}{cc}\# Significant \\ Slices Found*\end{tabular} & \begin{tabular}{cc}\# Candidate \\ Slices\end{tabular}  & \begin{tabular}{cc} CPU Usage \\ (CPU-sec)\end{tabular} & \begin{tabular}{cc} Shuffle \\ Bytes (GB)\end{tabular}\\
\midrule
\multirow{6}{*}{ \begin{tabular}{cc}OpenML \\ Electricity\end{tabular} } & Batch & 317 & 30000 & 168K & 233\\
                                    & Iterative & 327 & 23499 & 90.6K & 19.3 \\
                                    & Priority-Config 1 & 101 & 2859 & 14.1K & 0.967\\
                                    & Priority-Config 2  & 163 & 5638 & 17.8K & 2.72\\
                                    & Priority-Config 3 & 261 & 10724 & 40.3K & 8.36\\
                                    & Priority-Config 4 & 302 & 15114 & 54.9K & 9.68\\
\midrule
\multirow{6}{*}{\begin{tabular}{cc}UCI \\ Census\end{tabular}}  & Batch & 303 & 63861 & 412K & 971\\
                                    & Iterative & 316 & 29079 & 213K & 70.1\\
                                    & Priority-Config 1 & 195 & 5690 & 14.1K & 2.8\\
                                    & Priority-Config 2 & 258 & 10784 & 15.2K & 5.88\\
                                    & Priority-Config 3 & 321 & 24484 & 46.4K & 29.6\\
                                    & Priority-Config 4 & 374 & 30308 & 50.8K & 43.4\\
\midrule
\multirow{4}{*}{Traffic Stop**} 
                                    & Priority-Config 1 & 146 & 10157 & 1.34M & 326\\
                                    & Priority-Config 2 & 204 & 19317 & 1.79M & 981\\
                                    & Priority-Config 3 & 202 & 34442 & 1.44M & 1720\\
                                    & Priority-Config 4 & 245 & 65899 & 4.01M & 7350\\
\midrule                                    
\multirow{1}{*}{Safe Driving**} 
                                    & Priority-Config 1 & 369 & 10532 & 1.19M & 442\\
\bottomrule
\end{tabular}
\begin{tablenotes}
      \item * The number of significant slices may vary across runs due to the randomness of bootstrapping.
      \item ** The strategies or configurations that are not reported cannot finish in 8 hours. 
\end{tablenotes}
\end{table*}

\subsubsection{Micro-Benchmarks}\label{sec:micro_benchmark}
We provide several micro-benchmarks that validate the effectiveness of the design choices in \autoslicer.

\begin{figure}[t]
\begin{subfigure}{.45\linewidth}
  \centering
  \includegraphics[width=.9\linewidth]{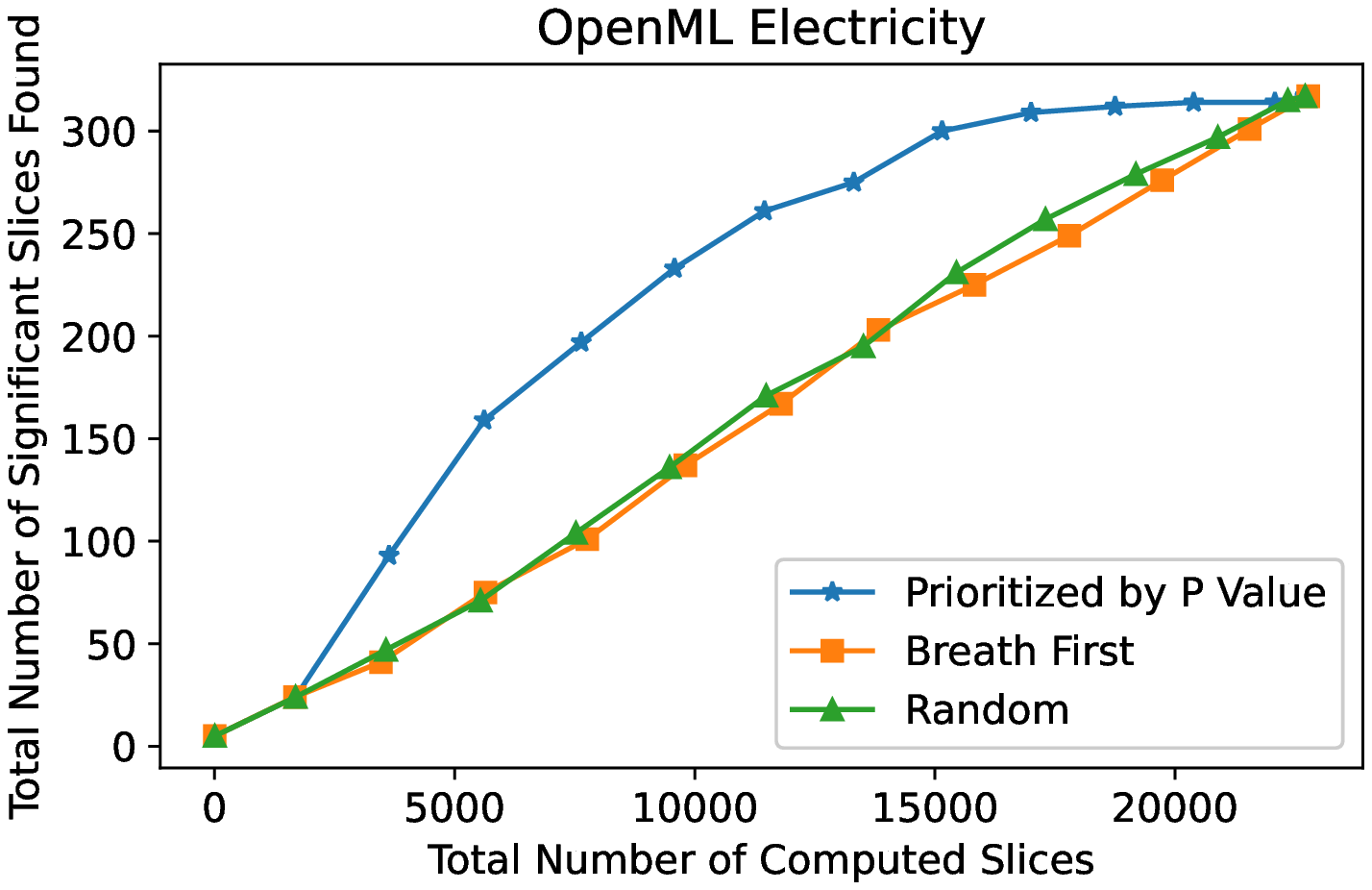}
\end{subfigure}%
\begin{subfigure}{.45\linewidth}
  \centering
  \includegraphics[width=.9\linewidth]{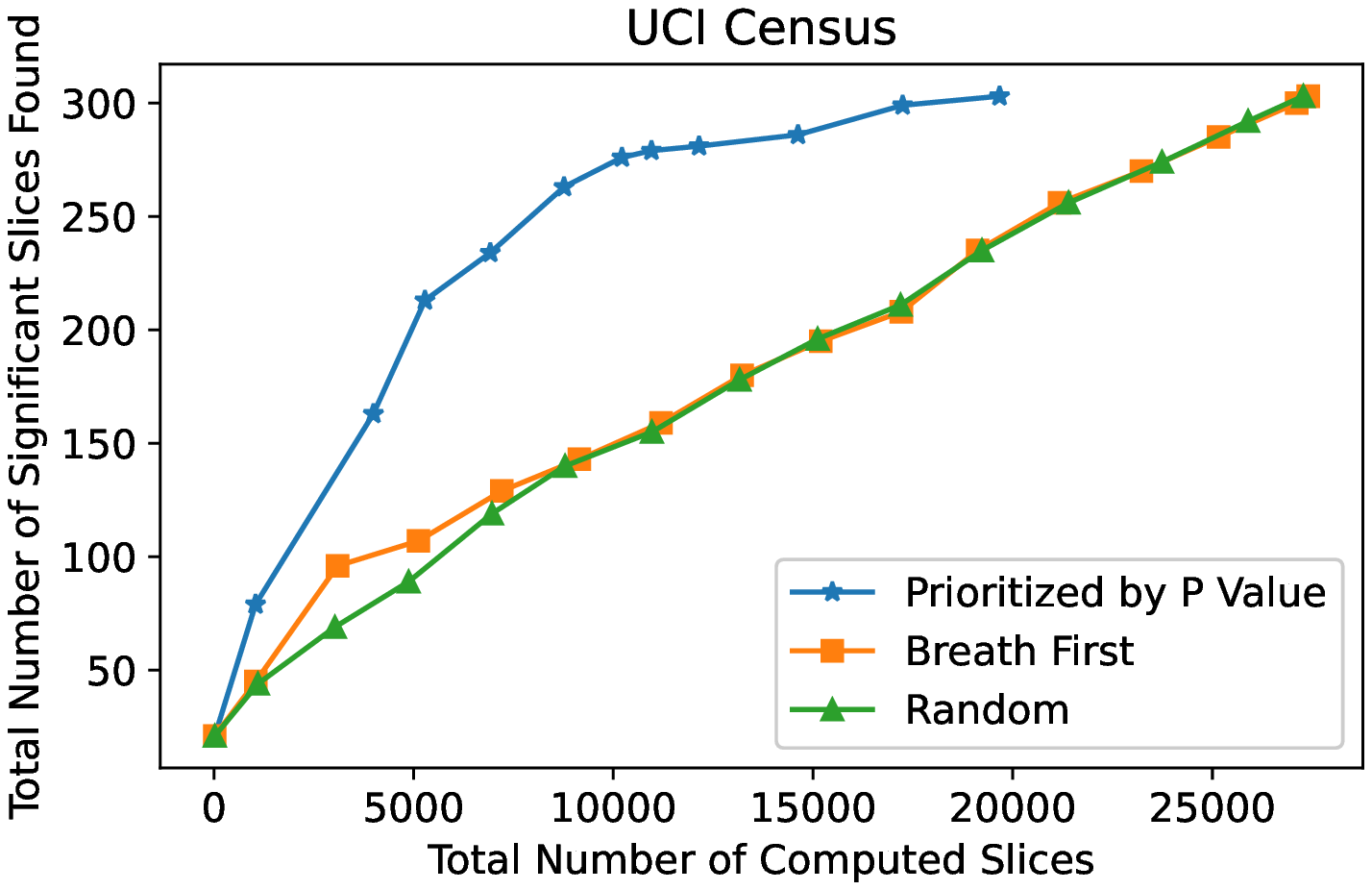}
\end{subfigure}
\caption{Number of significant slices found as the total number of computed slices varies.}
\label{fig:p_value_effectiveness}
\end{figure}

\newparagraph{Effectiveness of $p$-Value as Priority Score}
We compare different types of priority scores in the priority-queue-based iterative strategy. The results are shown in Figure~\ref{fig:p_value_effectiveness}. For the breath-first search, we use the depth (cross size) as the priority score. For the random search, we generate one random number for each slice as the priority score. We set the target number of candidate slices per iteration to $2000$, and report the total number of significant slices found as the actual number of computed candidates increases. The results show that \autoslicer\ can find more slices in its early stage with $p$-value as the priority score. For example, on OpenML Electricity, the $p$-value-prioritized strategy finds 90\% of the significant slices in its first $14000$ candidates, while the others take around $20000$ candidates. Similarly, on UCI Census, the $p$-value-prioritized strategy takes less than $12000$ candidates while the others take around $25000$ candidates.

\begin{figure}
\centering
\includegraphics[width=0.45\linewidth]{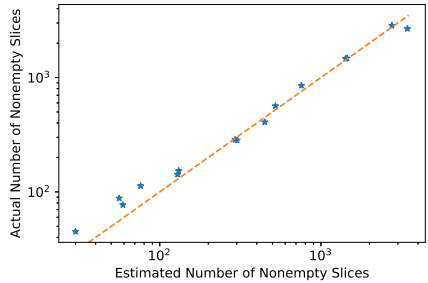}
\caption{Estimated and actual number of nonempty slices collected from different runs. The dashed line represents the ideal case where the actual number is the same as estimated.}
\label{fig:nonemptyEstimation}
\end{figure}

\newparagraph{Effectiveness of Nonempty Rate}
In the priority-queue-based iterative strategy , we keep track of the nonempty rate for each cross size to estimate the number of nonempty slices to be computed. We compare the estimated and the actual number of nonempty slices, and show the results in Figure~\ref{fig:nonemptyEstimation}. The data points are collected from several runs across different datasets. We can see that the estimated number of nonempty slices gets very close to the actual number by keeping track of the nonempty rate.

\newparagraph{Effectiveness of Minimum Slice Size}
\autoslicer\ uses the minimum slice size, $N_\text{min}$, to prune the search space for the iterative strategies. $N_\text{min}$ is set to 1 by default. Increasing $N_\text{min}$ significantly reduces the number of candidate slices to be explored. For example, when we run the simple iterative strategy on OpenML Electricity, the number of candidate slices explored reduces by 58.3\% when increasing $N_\text{min}$ to 50, and further reduces by 33.4\% when increasing $N_\text{min}$ to 100. 

\subsubsection{Scalability}\label{sec:scalability}
\begin{figure}
\centering
\includegraphics[width=0.45\linewidth]{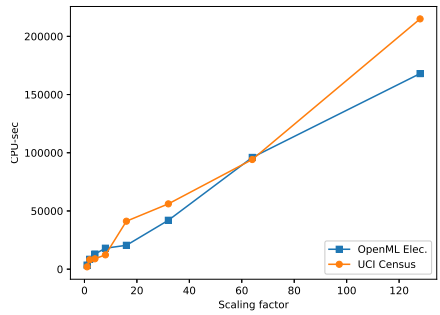}
\caption{CPU usage on increasing dataset size.}
\label{fig:scalability}
\end{figure}

We examine the scalability of \autoslicer\ by increasing the dataset size. Figure~\ref{fig:scalability} shows the CPU usage as we increase the dataset size. Here a value 4x on the x-axis means that we replicate the dataset 4 times. From the figure, we can see a linear relationship between the dataset size and the computation cost.

\subsection{Related Work}
\newparagraph{Data slicing for model debugging}
MLCube~\cite{kahng2016visual} and Manifold~\cite{zhang2018manifold} provide visualizing tools where the user can interactively specify slices and get their statistics. Slice-based Learning~\cite{chen2019slice} allows the user to specify slices to which extra model capacity will be allocated. \slicefinder~\cite{chung2019automated} searches for problematic slices where a model performs poorly in an automatic manner by statistical tests. It enumerates and computes one slice each time sequentially. \sliceline~\cite{sagadeeva2021sliceline} formulates slice finding as a linear-algebra problem, which can be solved efficiently by matrix computing tools.

\newparagraph{Other model debugging tools}
PALM~\cite{krishnan2017palm} approximates how much each training example contributes to the prediction so that the user can take actions for those that are responsible for the failure modes. MODE~\cite{ma2018mode} performs model state differential analysis to identify problematic internal features in the model and fix them by input selection and retraining. GEval~\cite{gralinski2019geval} finds problematic test data, errors in prepossessing and issues in models that contribute to performance degeneration for natural language processing tasks. Feature attribution~\cite{zhou2016learning, selvaraju2017grad, bau2017network, sundararajan2017axiomatic} techniques provide scores to quantify how each input feature contributes to the final prediction.

\newparagraph{Fairness in machine learning}
Machine learning fairness deals with the problem where a model treats certain groups of data in a biased way.
To solve this problem, some works~\cite{kamiran2012data, hajian2012methodology, calmon2017optimized} rely on the preprocessing steps to reduce bias in the training data. Another line of works~\cite{zafar2017fairness, agarwal2018reductions, celis2019classification, zhang2021omnifair} improve model fairness by incorporating fairness constraints in the training process. For example, OmniFair~\cite{zhang2021omnifair} allows the user to specify fairness constraints under which it maximizes the model accuracy. For the fairness problem, \autoslicer\ is a useful tool to identify groups that are treated unfairly by the model.

\newparagraph{Frequent Itemset Mining}
The problem of frequent itemset mining~\cite{agrawal1993mining} aims to find item sets that are frequently shown up together from a set of transactions. The lattice search space of automated slicing is similar to that of the frequent itemset mining if we treat each singleton predicate as a single item and the conjunction of them as itemsets. Algorithms solving the frequent itemset mining problem~\cite{agrawal1994fast, han2000mining} typically prune the search space based on the downward closure lemma which states that subsets of a frequent itemset must also be frequent. We also rely on this property to prune the candidates whose parent contains less than $N_\text{min}$ examples. However, such monotonicity property does not hold for the significance of slices in the automated slicing problem.

\end{document}